\documentclass{article}

\usepackage[round]{natbib}

\usepackage[main, final]{neurips_2025}

\usepackage[utf8]{inputenc} 
\usepackage[T1]{fontenc}    
\usepackage{hyperref}       
\usepackage{url}            
\usepackage{booktabs}       
\usepackage{amsfonts}       
\usepackage{nicefrac}       
\usepackage{microtype}      
\usepackage{xcolor}         
\usepackage{enumitem}       
\usepackage{amsmath}
\usepackage{svg} 
\usepackage{algorithm}
\usepackage{algpseudocode}

\usepackage{booktabs}   
\usepackage{multirow}   
\usepackage{array}      

\title{FMBench: Adaptive Large Language \\ Model Output Formatting}

\author{%
  Yaoting Wang, Yun Zhou, Henghui Ding \\
  Fudan University, China \\
}

\begin{document}

\maketitle

\begin{abstract}
  Producing outputs that satisfy both \,\emph{semantic intent} and \,\emph{format constraints} is essential for deploying large language models in user-facing and system-integrated workflows. In this work, we focus on Markdown formatting, which is ubiquitous in assistants, documentation, and tool-augmented pipelines but still prone to subtle, hard-to-detect errors (e.g., broken lists, malformed tables, inconsistent headings, and invalid code blocks) that can significantly degrade downstream usability.
  We present \textbf{FMBench}, a benchmark for adaptive Markdown output formatting that evaluates models under a wide range of instruction-following scenarios with diverse structural requirements. FMBench emphasizes real-world formatting behaviors such as multi-level organization, mixed content (natural language interleaved with lists/tables/code), and strict adherence to user-specified layout constraints.
  To improve Markdown compliance without relying on hard decoding constraints, we propose a lightweight alignment pipeline that combines supervised fine-tuning (SFT) with reinforcement learning fine-tuning. Starting from a base model, we first perform SFT on instruction--response pairs, and then optimize a composite objective that balances semantic fidelity with structural correctness.
  Experiments on two model families (OpenPangu and Qwen) show that SFT consistently improves semantic alignment, while reinforcement learning provides additional gains in robustness to challenging Markdown instructions when initialized from a strong SFT policy. Our results also reveal an inherent trade-off between semantic and structural objectives, highlighting the importance of carefully designed rewards for reliable formatted generation. Code is available at: \url{https://github.com/FudanCVL/FMBench}.
\end{abstract}

\section{Introduction}
Large language models (LLMs)~\citep{chen2025pangu,achiam2023gpt,liu2024deepseek,yang2025qwen3} have rapidly become the default interface for knowledge access and task automation. At the same time, modern alignment pipelines, from instruction tuning to RLHF-style preference optimization, have substantially improved their ability to follow natural language instructions and produce responses that appear helpful at first glance~\citep{ouyang2022training}. Yet, in many production settings, success is not determined solely by \emph{what} the model says: it is equally shaped by \emph{how} the answer is packaged and how it will be consumed. In particular, a response must be readable to humans, reliably parseable by downstream tools, and consistent with user-specified presentation requirements.

In this work we focus on \textbf{Markdown-constrained generation}. Markdown is the de facto interchange format for chat assistants, technical documentation, and many tool-augmented workflows because it is both human-friendly and structurally expressive: it can be rendered consistently across clients, inspected by users, and post-processed by lightweight parsers when needed. At the same time, Markdown formatting is notoriously error-prone for LLMs. Models often produce broken nested lists, malformed tables, inconsistent heading hierarchies, or unbalanced fenced code blocks, even when the underlying content is largely correct, and these failures are ``small'' at the token level but disproportionately harmful in practice.

A key challenge is that Markdown constraints are frequently \emph{soft} rather than purely syntactic. Even when multiple renderings are technically valid, only some match human conventions or user intent (e.g., consistent heading levels, stable ordering of sections, or clear separation between prose and code). Consequently, ``format compliance'' is not binary: an output can be technically well-formed yet still hard to read, inconsistent across sections, or misaligned with the organizational structure implied by the instruction. Moreover, format restrictions can interact with task performance and affect downstream correctness~\citep{tam2024let}, motivating an increasing emphasis on reliable structured outputs and templated interaction patterns~\citep{shen2025slot} as well as benchmarks for structured generation and instruction following~\citep{yang2025structeval,ye2025multi}.

How should we make Markdown outputs reliable? One direction is to enforce validity at inference time via constrained decoding or grammar-based generation~\citep{beurer2024guiding,dong2025xgrammar}, which can prevent certain classes of structural errors by restricting allowable continuations. However, strict constraints may introduce inference overhead and do not by themselves ensure \emph{human-preferred} formatting, especially in Markdown where usability depends on global organization and cross-section consistency. A complementary direction is to internalize formatting behavior through post-training; for instance, format-aware reinforcement learning has been explored to improve format faithfulness across tasks~\citep{yao2025reff}. In principle, this can improve robustness without external constraints at inference time, but it still depends on having appropriate evaluation signals for both content and format.

Despite this progress, two gaps remain. \textbf{First}, existing benchmarks~\citep{yang2025structeval,ye2025multi} rarely isolate the Markdown setting that dominates assistant usage, making it difficult to diagnose and compare methods under realistic Markdown requirements. \textbf{Second}, the objective is inherently multi-dimensional: optimizing for semantic correctness alone does not guarantee format compliance, while aggressively optimizing for structure can distort content. This motivates both a focused benchmark and a training recipe that explicitly balances these competing goals.

We address these gaps with \textbf{FMBench}, a benchmark that focuses exclusively on \emph{Markdown} output formatting. FMBench spans a diverse set of formatting requirements, including hierarchical sections, mixed prose--list--table composition, and code blocks, and evaluates outputs along two complementary axes: semantic correctness and Markdown structural compliance. Building on FMBench, we propose an alignment pipeline that combines SFT with reinforcement learning fine-tuning (RLFT) to optimize both objectives without requiring hard constraints at inference time.

This paper makes the following contributions:
\begin{itemize}[
      leftmargin=*,
      topsep=2pt,
      itemsep=4pt,
      parsep=0pt
    ]
  \item \textbf{Benchmark:} We introduce \textbf{FMBench}, a Markdown-focused benchmark for evaluating adaptive output formatting under realistic instruction-following settings.
  \item \textbf{Method:} We present a practical post-training recipe from SFT to RL that improves Markdown compliance while maintaining semantic fidelity.
  \item \textbf{Analysis:} We provide empirical results across two model families (OpenPangu and Qwen) and analyze the trade-off between semantic and structural objectives in Markdown generation.
\end{itemize}

\section{Related Work}

\subsection{Formatting Outputs of LLMs}

In many real-world deployments of LLMs, it is essential that model outputs strictly conform to predefined formats. Such requirements arise in a wide range of applications, including structured data generation, form filling, code synthesis, and API interaction governed by rigid protocols. Even minor formatting violations may trigger parsing failures, disrupt downstream pipelines, or severely degrade system reliability. As a result, enabling LLMs to consistently produce outputs that satisfy both syntactic and semantic constraints has become an important problem studied across both academia and industry.
A widely adopted strategy is prompt engineering~\citep{mao2025prompts,he2024does,liu2025beyond}, which guides models toward desired output structures by embedding explicit instructions, demonstrations, or templates in the input prompt. While prompt-based methods are attractive due to their simplicity and zero-training cost, their effectiveness is highly sensitive to prompt wording, context organization, and the model’s intrinsic instruction-following capability. Small variations in prompt phrasing or input distribution can lead to disproportionate formatting errors, resulting in poor robustness and limited reproducibility in practical settings.
Another line of work focuses on constrained decoding~\citep{beurer2024guiding,dong2025xgrammar}, where the token generation space is dynamically restricted during inference so that only grammar-compliant tokens are permitted at each step. This approach offers strong guarantees of syntactic correctness and is particularly effective for formats that admit precise formal specifications, such as regular languages or context-free grammars. However, constrained decoding typically incurs substantial computational overhead due to the need for grammar state tracking or real-time validation, often leading to slower inference. Moreover, when the imposed structural constraints are misaligned with the model’s subword tokenization or natural language generation dynamics, hard constraints may adversely affect semantic fluency or content quality. This issue is especially pronounced in hybrid outputs that interleave natural language with structured fields.
Post-training fine-tuning approaches have also been explored to improve structural adherence, achieving promising results in tasks such as named entity recognition~\citep{wang2025gpt} and information extraction~\citep{wang2023instructuie}. By exposing models to large volumes of format-specific training data, these methods enable models to internalize target output structures. Nevertheless, in formatting-oriented scenarios, such approaches often overfit to superficial regularities in the training distribution, such as fixed field orderings or specific punctuation patterns, rather than learning robust, context-aware formatting principles. Consequently, their generalization performance degrades when encountering unseen format variants or slight perturbations in user instructions.

\subsection{Reinforcement Learning for Output Formatting}
Reinforcement learning (RL) provides a general framework for learning decision-making policies through interaction with an environment and feedback in the form of reward signals. Unlike supervised learning, RL does not rely on explicit labeled targets and is therefore well suited to optimization problems involving implicit, multi-dimensional constraints. In the context of LLMs, RL has emerged as a key technique for improving instruction following, alignment, and controllability, especially when desired behaviors cannot be easily specified via static supervision.
Conventional SFT often struggles to capture the nuanced constraints embedded in complex instructions, while reinforcement learning from human feedback (RLHF)~\citep{ouyang2022training}, despite its success in aligning LLMs with human preferences, suffers from several limitations. These include the ambiguity and subjectivity of human-provided rewards, high annotation costs, and limited sample efficiency, particularly for fine-grained constraints such as output formatting correctness.
To address these challenges, recent work has increasingly focused on verifiable reward design, which replaces or augments subjective human judgments with automatically evaluable signals derived from formal rules, programmatic checkers, or structured rubrics. ReFF~\citep{yao2025reff} employs programmatic format validators to generate reinforcement signals, significantly improving structural compliance without sacrificing content quality. VERIF~\citep{peng2025verif} further enhances reward reliability by integrating rule-based code verification with LLM-assisted verification. HiR~\citep{zhang2025replay} explores joint optimization of semantic fidelity and structural correctness through dynamic format exploration strategies. RIFL~\citep{he2025advancedif} introduces rubric-driven reward shaping to handle multiple, potentially competing instruction constraints, while RLCF~\citep{viswanathan2025checklists} derives checklist-based rewards to support scalable and interpretable reinforcement learning.
Collectively, these studies demonstrate the effectiveness of reinforcement learning with verifiable rewards in improving instruction adherence and structural correctness. However, existing approaches often rely on task-specific reward engineering or rigid rule definitions, which may limit adaptability across diverse formatting schemas or evolving task requirements. This leaves open the question of how to design reinforcement signals that are both verifiable and sufficiently flexible to generalize across heterogeneous formatting constraints.

\section{FMBench Dataset}

\label{sec:data_pipeline}

\begin{figure}[ht]
  \centering
  \includegraphics[width=1\textwidth]{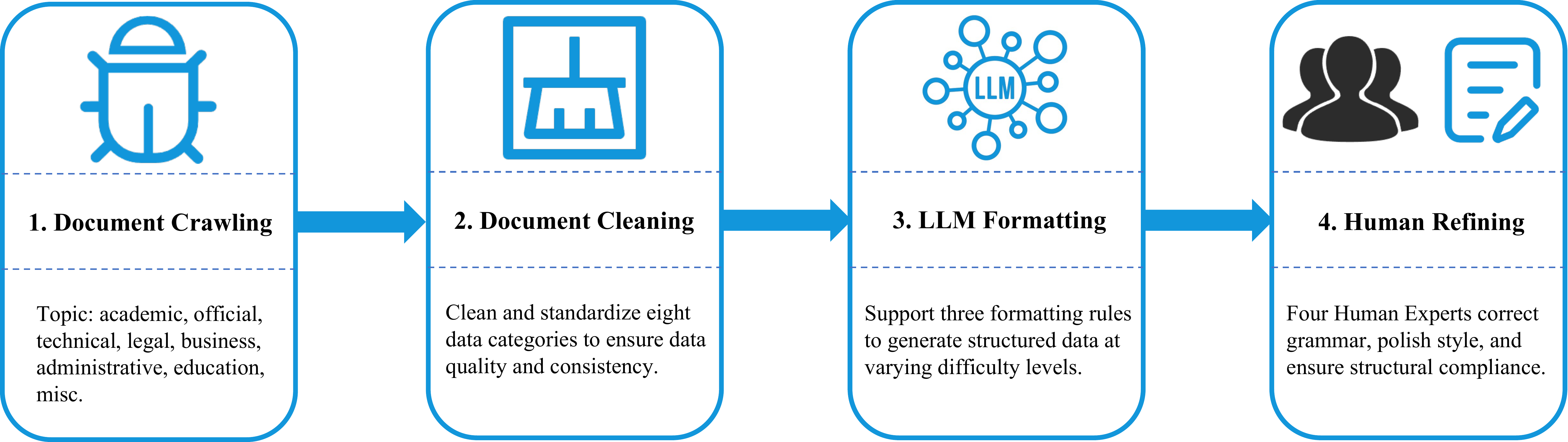}
  \caption{Overview of a four-stage pipeline for converting raw documents into structured data.
(1) Document Crawling: Documents are collected from multiple domains, including academic, official, technical, legal, business, and educational sources.
(2) Document Cleaning: Documents are cleaned and standardized across eight data categories to ensure quality and consistency.
(3) LLM Formatting: LLMs apply three formatting rules to produce structured data at varying difficulty levels.
(4) Human Refining: Human experts review and refine the outputs to correct grammar, improve style, and ensure structural compliance.}
  \label{fig:dataset_pipeline}
\end{figure}

\subsection{Data Pipeline}

We develop a semi-automated data construction pipeline for structured document synthesis under a content-preserving structural transformation setting. In this setting, models are required to reorganize documents according to explicit formatting specifications while strictly preserving the original content and its ordering. The pipeline is designed to generate diverse, high-quality, and strictly formatted examples, while maintaining full reproducibility and auditability.

\subsubsection{Pipeline Stages}

\paragraph{Input Collection and Preprocessing.}
The pipeline begins by collecting raw seed documents spanning eight representative document types, including academic papers and official documents, which serve as structural and stylistic templates. Each input example is ingested as a JSONL record containing a raw text field; empty or malformed entries are discarded at ingestion time. The remaining documents are processed independently, enabling embarrassingly parallel generation across seeds and variants.
Prior to structural transformation, the raw texts are cleaned using automated rules that normalize formatting inconsistencies, remove noise, and repair structural defects such as missing section boundaries or malformed lists.

\paragraph{Atomic Unit Segmentation.}
Each cleaned document is segmented into ordered \emph{atomic content units} to prevent uncontrolled changes.
Rule-based segmentation separates headings, metadata fields, and paragraph sentences using conservative sentence splitting.
Atomic units are indivisible and cannot be reordered, merged, or split in subsequent stages.

\paragraph{Specification Sampling and Structure Generation.}
For each seed, the pipeline deterministically samples a Markdown structural specification that defines section hierarchy, permitted block types (e.g., lists, tables, block quotes, code blocks), and formatting constraints such as line wrapping. We expose discrete difficulty levels that control structural complexity, with higher levels enforcing deeper hierarchies and stricter formatting requirements. All stochastic choices are derived from a hash-based pseudo-random generator seeded by the pair of seed index and variant index, ensuring full reproducibility.

\paragraph{Structure-Constrained Construction.}
Given the segmented atomic units and a sampled specification, the pipeline constructs the target Markdown document by sequentially assigning content units to instantiated structural blocks. Content units are consumed strictly in their original order. Metadata-style field lines, when present, are grouped into a dedicated \texttt{Metadata} section. When a structural block requires more content slots than available units, deterministic padding rules are applied to preserve structural validity without introducing new semantic content.

\subsubsection{Quality Control and Verification}
To further improve linguistic quality and formatting consistency, we apply a LLM to perform semantic refinement and format standardization on the automatically generated documents. The resulting outputs are then reviewed through a human-in-the-loop verification process, where annotators correct residual formatting errors and filter out low-quality samples. This hybrid design balances scalability with quality assurance.

\vspace{-2mm}
\subsubsection{Outputs and Metadata}
For each generated variant, we render a canonical instruction prompt that specifies hard constraints, including Markdown-only output, strict content preservation, and order invariance, together with the sampled structural specification. Each dataset entry includes the original seed text, the instruction prompt, the target Markdown document, the specification, and auxiliary metadata such as difficulty level and a versioned validator identifier. This design enables downstream automatic verification and fine-grained evaluation of structure adherence.

\vspace{-2mm}
\subsection{Dataset Statistics.}
\begin{figure}[h]
  \centering
  \includegraphics[width=1\textwidth]{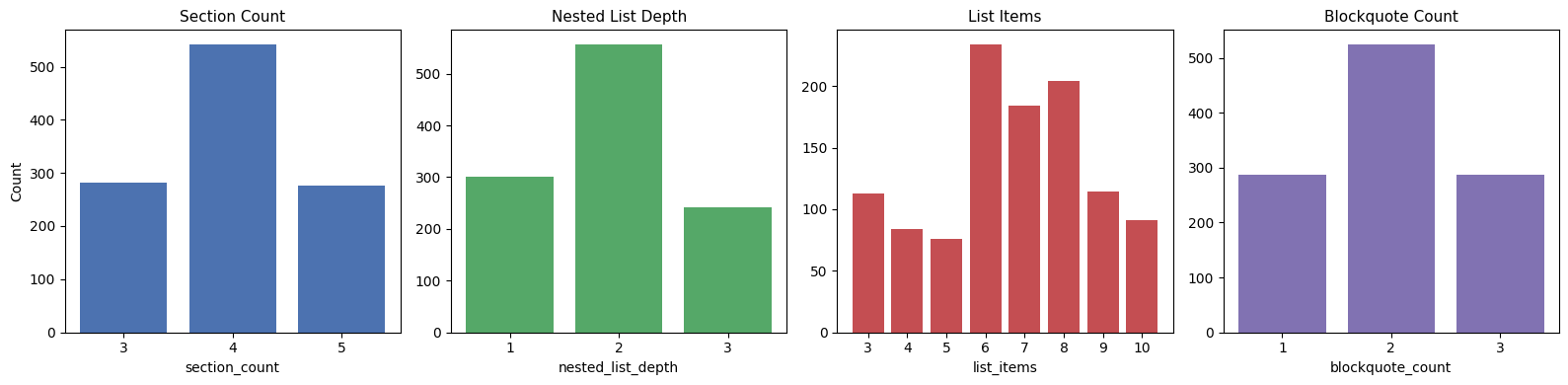}
  \caption{Distribution of key structural parameters in the FMBench dataset.
    The figure shows the marginal distributions of four core structural variables:
    section count, nested list depth, number of list items, and blockquote count,
    computed over the combined training and test sets. The distributions reveal a
    highly controlled design with symmetric or unimodal patterns, where structural
    complexity is primarily modulated by list-related constraints rather than by
  section or blockquote counts.}
  \label{fig:stat}
\end{figure}

Following this pipeline, we construct \textbf{FMBench}, a dataset comprising 1,100 high-quality Markdown-formatted documents. We split the dataset into 800 training samples and 300 test samples, enabling reliable evaluation of formatting generalization across document types and structural complexities.
In addition, as illustrated in Figure~\ref{fig:stat}, the FMBench dataset exhibits a carefully balanced and constrained structural design. The number of sections and blockquotes follows nearly symmetric distributions centered around moderate values, indicating that these variables primarily define a stable document scaffold rather than serving as major sources of difficulty. In contrast, the depth of nested lists and the number of list items display broader and more varied distributions, suggesting that they constitute the main axes of structural complexity in the dataset. This design choice concentrates generation difficulty on maintaining hierarchical consistency and list integrity under moderate structural pressure, while avoiding extreme or unnatural configurations. Consequently, the dataset emphasizes robustness to structural constraints over raw scale or depth, aligning the task more closely with realistic long-form formatting scenarios.

\subsection{Evaluation Metrics}

We report two quantitative metrics on the FMBench test set:
(a) \textbf{Semantic Score}, measured by BERTScore-F1~\citep{zhang2019bertscore} between the generated output and the reference text, reflecting semantic preservation.
And (b) \textbf{Structure Score}, measured by the structural reward described above, reflecting compliance with the target formatting and document organization.
Both metrics are averaged over the full test set.

\section{Experiments}

\subsection{Experimental Setup}

\begin{figure}[h]
  \centering
  \includegraphics[width=1\textwidth]{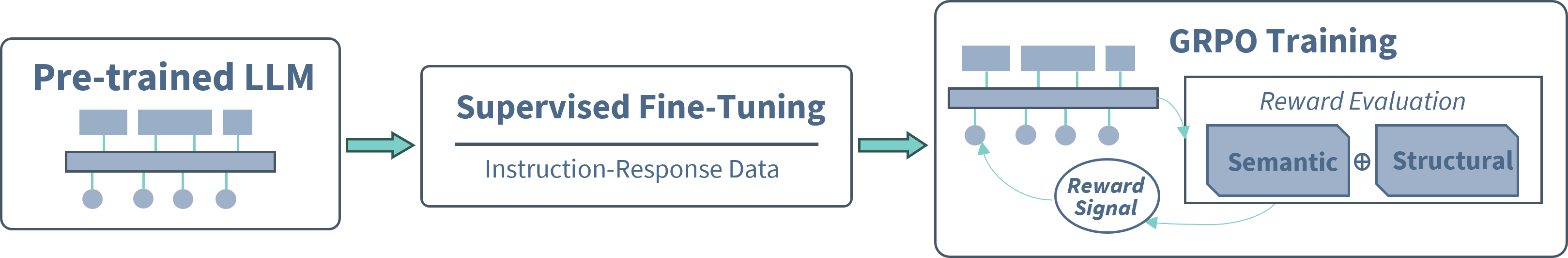}
  \caption{Training pipeline.}
  \label{fig:llm}
\end{figure}

We evaluate our approach on the FMBench benchmark, which focuses on content-preserving formatted text generation. All experiments are conducted on two representative model families with different scales: OpenPangu (1B and 7B) ~\citep{chen2025pangu} and Qwen3 (1.7B and 8B)~\citep{yang2025qwen3}. For each model, as shown in Figure~\ref{fig:llm}, we consider four training configurations: (i) the original pretrained model, (ii) supervised fine-tuning (SFT), (iii) reinforcement learning fine-tuning with GRPO~\citep{shao2024deepseekmath} initialized from the pretrained model, and (iv) SFT followed by GRPO (SFT+GRPO), which constitutes our full method.
Unless otherwise specified, all models are evaluated on the same held-out test split of FMBench using identical decoding settings. We fix the maximum generation length and decoding strategy across all methods to ensure fair comparison.
All inputs are formatted using each model’s native chat template and truncated to a maximum sequence length of 2{,}048 tokens. We compute a completion-only loss during training, restricting gradient updates to tokens belonging to the assistant’s response segment. Sequence packing is disabled, as documents in FMBench typically approach the maximum length. All training is conducted in bfloat16 precision on 8 Ascend NPUs using the AdamW optimizer.

\subsection{Supervised Fine-Tuning}

In the SFT stage, we adapt the base models to the formatted text generation task using parameter-efficient fine-tuning. Specifically, we fine-tune OpenPangu-1B/7B and Qwen3-1.7B/8B for one epoch using Low-Rank Adaptation (LoRA)~\citep{hu2022lora}.
We employ a constant learning rate of $2 \times 10^{-5}$ with a batch size of 1 and a weight decay of 0.01. For the LoRA configuration, we set the rank to 8 with a scaling factor $\alpha = 64$ and apply 10\% dropout for regularization. Adaptation is applied exclusively to the query and value projection matrices in the self-attention layers, following common practices to minimize parameter overhead while preserving attention expressiveness.

\subsection{Reinforcement Learning with Verifiable Rewards}

Following SFT, we apply reinforcement learning fine-tuning using GRPO to further improve structural compliance while preserving semantic fidelity. GRPO is initialized either from the pretrained model or from the SFT checkpoint, depending on the experimental setting.
For each input prompt, we sample four parallel candidate generations and optimize the policy via group-wise relative reward comparisons. The overall reward is defined as
\[
  r = \lambda_1 r_{\text{sem}} + \lambda_2 r_{\text{struct}},
\]
where $\lambda_1 = \lambda_2 = 1$.
The semantic reward $r_{\text{sem}}$ is computed using BERTScore-F1 between the generated output and the reference. To better handle long documents, we employ a Longformer-base-4096 encoder and extract representations from layer 5. The structural reward $r_{\text{struct}}$ targets higher-level organizational similarity: we first generate abstractive summaries of both the generated output and the reference using a frozen language model, and then compute their BERTScore similarity. This design encourages alignment at the discourse and structural level while remaining robust to lexical variation. We do not apply explicit reward normalization, as GRPO is empirically robust to reward scale differences due to its relative ranking mechanism.

\subsection{Main Results Analysis}

Table~\ref{tab:main_results} presents the main results on FMBench across different model families and scales. 
We compare the pretrained baseline with our full training pipeline, i.e., supervised fine-tuning followed by GRPO (SFT+GRPO).
Overall, applying SFT+GRPO consistently improves semantic fidelity over the pretrained models while preserving strong structural compliance.
This trend holds across all evaluated architectures, indicating that our training pipeline generalizes well across both model families and parameter scales.

For smaller models such as OpenPangu-1B and Qwen3-1.7B, SFT+GRPO yields clear semantic gains compared to the pretrained baseline, while maintaining comparable structural performance.
Notably, for larger models (e.g., Qwen3-8B), SFT+GRPO achieves the best overall semantic score together with the highest structural compliance.
This suggests that larger-capacity models are better able to jointly optimize semantic fidelity and structural constraints under our training strategy.

\begin{table}[ht]
  \centering
  \begin{tabular}{lccc}
    \toprule
    \textbf{Model} & \textbf{Pretrained} & \textbf{SFT+GRPO} & \textbf{Metric} \\
    \midrule
    OpenPangu-1B & 0.9300 & 0.9482 & Semantic \\
                 & 0.9535 & 0.9603 & Structure \\
    \midrule
    OpenPangu-7B & 0.9318 & 0.9466 & Semantic \\
                 & 0.9712 & 0.9710 & Structure \\
    \midrule
    Qwen3-1.7B   & 0.9395 & 0.9467 & Semantic \\
                 & 0.9652 & 0.9652 & Structure \\
    \midrule
    Qwen3-8B     & 0.9347 & 0.9507 & Semantic \\
                 & 0.9700 & 0.9708 & Structure \\
    \bottomrule
  \end{tabular}
  \vspace{3mm}
  \caption{Main results on FMBench. We compare pretrained models with the full training pipeline (SFT+GRPO), reporting semantic and structural scores.}
  \label{tab:main_results}
\end{table}

\subsection{Ablation Study}

To analyze the contribution of each component in our training pipeline, we conduct an ablation study by progressively adding SFT and GRPO to the pretrained models.
The detailed results are shown in Table~\ref{tab:ablation}.

Across all model families and scales, introducing SFT leads to substantial improvements in semantic fidelity compared to the pretrained baseline.
In contrast, structural performance remains largely unchanged after SFT, indicating that supervised alignment primarily enhances semantic correctness without degrading format compliance.

When GRPO is applied on top of SFT, structural scores are consistently improved or better preserved, while semantic performance is largely maintained.
For smaller models, GRPO may slightly trade off semantic accuracy in favor of stronger structural consistency.
However, for larger models such as Qwen3-8B, the combination of SFT and GRPO achieves the best semantic performance overall, suggesting that sufficient model capacity is crucial for effectively balancing semantic and structural objectives during reinforcement-style optimization.

Overall, these results indicate that SFT is the dominant factor driving semantic improvements on FMBench, while GRPO serves as a complementary component that reinforces structural compliance, particularly in higher-capacity models.

\begin{table}[ht]
  \centering
  \begin{tabular}{lccccc}
    \toprule
    \textbf{Model} & \textbf{Pretrained} & \textbf{+SFT} & \textbf{+GRPO} & \textbf{+SFT+GRPO} & \textbf{Metric} \\
    \midrule
    OpenPangu-1B & 0.9300 & \textbf{0.9601} & 0.9308 & 0.9482 & Semantic \\
                 & 0.9535 & 0.9602 & 0.9528 & \textbf{0.9603} & Structure \\
    \midrule
    OpenPangu-7B & 0.9318 & \textbf{0.9481} & 0.9319 & 0.9466 & Semantic \\
                 & 0.9712 & 0.9705 & 0.9717 & \textbf{0.9710} & Structure \\
    \midrule
    Qwen3-1.7B   & 0.9395 & \textbf{0.9480} & 0.9409 & 0.9467 & Semantic \\
                 & 0.9652 & \textbf{0.9660} & 0.9644 & 0.9652 & Structure \\
    \midrule
    Qwen3-8B     & 0.9347 & 0.9477 & 0.9346 & \textbf{0.9507} & Semantic \\
                 & 0.9700 & 0.9704 & 0.9693 & \textbf{0.9708} & Structure \\
    \bottomrule
  \end{tabular}
  \vspace{3mm}
  \caption{Ablation study on FMBench. We progressively add SFT and GRPO to pretrained models and report semantic and structural scores.}
  \label{tab:ablation}
\end{table}

\section{Discussion and Future Directions}
Our study underscores that reliable LLM deployment in tool-facing and user-facing workflows depends not only on semantic correctness but also on stable, human-usable formatting. By isolating Markdown as a first-class target format, FMBench reveals a persistent gap between answers that are ``mostly correct'' in content and outputs that are robustly renderable and parseable. More broadly, our results suggest that formatting behavior should be treated as an explicit alignment objective rather than a byproduct of instruction following.

\paragraph{Limitations.}
First, although FMBench spans a wide range of Markdown structures, it cannot cover the full long tail of real-world authoring styles and renderer-specific behaviors. Some choices are inherently subjective (e.g., when to split sections, how verbose headings should be), and any structural metric will encode particular conventions. Second, our evaluation is largely reference-based; in practice, there may exist multiple equally acceptable Markdown realizations for the same intent, which a single target cannot fully capture. Third, the proposed reward design depends on pretrained encoders and summarizers, which may introduce bias and may miss certain fine-grained formatting errors (e.g., subtle table misalignment or whitespace-sensitive issues) that are critical for specific downstream parsers.

\paragraph{Implications for reward design.}
The observed semantic--structure tension indicates that na\"ively increasing structural reward weight can encourage over-regularized, template-like outputs that inadvertently omit or distort requested content. A promising direction is to move from purely similarity-based structure rewards toward validator-grounded and instruction-conditioned rewards: executable checks (e.g., code-fence balance, list nesting validity, table column consistency) can provide crisp signals, while conditioning on the requested layout can better distinguish ``valid but wrong'' structures from genuinely compliant formatting.

\paragraph{Future directions.}
We see several concrete opportunities to extend this work. (1) \textbf{Beyond Markdown:} expanding from Markdown to other structured outputs commonly used in tool calls (e.g., JSON/YAML), as well as hybrid settings (e.g., Markdown with embedded schema blocks). (2) \textbf{Hybrid constraint + learning:} combining post-training with lightweight inference-time validation or constrained decoding to guarantee syntactic validity while preserving flexible, human-preferred organization. (3) \textbf{Robustness and transfer:} evaluating cross-domain generalization (documentation, customer support, agent tool-use) and stress-testing with adversarial instructions that trigger subtle structural failures. (4) \textbf{Human-centered evaluation:} complementing automatic scores with user studies that measure readability, trust, and downstream task success when outputs are consumed by both humans and parsers.

Overall, we hope FMBench can serve as a practical diagnostic tool to motivate further work on alignment objectives that treat formatting as a first-class reliability requirement.

\section{Conclusion}
We introduced \textbf{FMBench}, a benchmark that targets adaptive Markdown output formatting, and presented a lightweight post-training pipeline that combines SFT with GRPO to jointly improve semantic fidelity and structural compliance. Experiments on two model families (OpenPangu and Qwen) show that SFT consistently strengthens semantic alignment, while adding GRPO further improves or better preserves structural robustness, especially when initialized from a strong SFT policy.
Beyond aggregate improvements, our analysis reveals an inherent trade-off between semantic and structural objectives, highlighting the importance of reward designs that are both validator-grounded and instruction-conditioned. We hope FMBench will serve as a practical diagnostic tool for formatting reliability, and will catalyze future research on alignment methods that treat formatting as a first-class requirement for dependable structured generation.

\newpage
\bibliographystyle{plainnat}
\bibliography{Ref}

\end{document}